\documentclass{article}

\usepackage{microtype}
\usepackage{graphicx}
\usepackage{subcaption}
\usepackage{booktabs} 
\usepackage{tabularx}
\newcolumntype{Y}{>{\raggedright\arraybackslash}X}

\usepackage{hyperref}

\usepackage[preprint]{icml2026}

\usepackage{amsmath}
\usepackage{amssymb}
\usepackage{mathtools}
\usepackage{amsthm}
\usepackage{enumitem}

\usepackage[capitalize,noabbrev]{cleveref}

\theoremstyle{plain}

\theoremstyle{definition}

\theoremstyle{remark}

\usepackage[textsize=tiny]{todonotes}

\icmltitlerunning{XForecast: The KDD 2026 Finance Day Challenge on News-Grounded Financial Forecasting}

\begin{document}

\twocolumn[
  \icmltitle{XForecast: The KDD 2026 Finance Day Challenge \\ on News-Grounded Financial Forecasting}



  \icmlsetsymbol{equal}{*}

  \begin{icmlauthorlist}
    \icmlauthor{Jaehoon Lee}{lg}
    \icmlauthor{Jun Seo}{lg}
    \icmlauthor{Seunghan Lee}{lg}
    \icmlauthor{Tae Yoon Lim}{lg}
    \icmlauthor{Dongwan Kang}{lg}
    \icmlauthor{Hwanil Choi}{lg}
    \icmlauthor{Minjae Kim}{lg}
    \icmlauthor{Sungdong Yoo}{lg}
    \icmlauthor{Junhyeok Kang}{lg}
    \icmlauthor{Sangjun Han}{lg}
    \icmlauthor{Soonyoung Lee}{lg}
    \icmlauthor{Yongjae Lee}{unist}
    \icmlauthor{Wonbin Ahn}{lg}
  \end{icmlauthorlist}

  \icmlaffiliation{lg}{LG AI Research, Seoul, South Korea}
  \icmlaffiliation{unist}{Ulsan National Institute of Science and Technology (UNIST), Ulsan, South Korea}

  \icmlcorrespondingauthor{Wonbin Ahn}{wonbin.ahn@lgresearch.ai}

  \icmlkeywords{Machine Learning, KDD, Finance}

  \vskip 0.3in
]



\printAffiliationsAndNotice{}  

\begin{abstract}
  Financial forecasting has traditionally relied on numerical time-series (TS) signals, while textual information such as news remains largely underutilized. We introduce XForecast, a benchmark and competition for news-grounded financial forecasting, where the task is to predict stock price movements four weeks ahead using both historical price data and paired news signals. XForecast combines stock price data with curated financial text generated using FinTexTS, a multimodal dataset generation method. It covers 100 large-cap stocks from 2019--2023, with training on 2019--2022 and evaluation on 2023. To better evaluate models that leverage textual signals, we adopt a weighted hit rate metric that places greater emphasis on instances where news information is expected to have stronger influence. The XForecast Challenge, hosted on Kaggle as part of KDD 2026 Finance Day, aims to benchmark methods for integrating financial TS and textual information. The competition is publicly available at:
\url{https://www.kaggle.com/competitions/xforecast-challenge-kdd}.
\end{abstract}

\section{Introduction}

Financial forecasting has long been a central problem in quantitative finance and machine learning. Most existing approaches primarily rely on numerical time-series (TS) signals, such as historical prices, trading volumes, and technical indicators~\cite{GIANTSIDI2025104719,yan2024doublepathadaptivecorrelationspatialtemporalinverted,zhao2024doubleadaptmetalearningapproachincremental,soroka2026dataefficient}. While these signals capture certain market dynamics, textual information such as financial news remains largely underutilized, despite its significant role in reflecting market-relevant events and information.

Recent advances in natural language processing and representation learning have made it increasingly feasible to incorporate textual information into financial prediction models. However, progress in this direction remains limited due to the lack of standardized benchmarks that systematically integrate financial TS data with aligned textual signals. Many existing datasets focus primarily on sentiment analysis or provide loosely associated news streams, making it difficult to rigorously evaluate how effectively models leverage textual information for financial forecasting tasks.~\cite{liu2024timemmd,xie2024fnspid}

To address this limitation, we construct a dataset that aligns financial TS data with curated financial text. In particular, we leverage FinTexTS~\cite{lee2026fintexts}, a multimodal dataset generation method that produces multi-level paired financial texts capturing macroeconomic, sector-level, related-company, and target-company contexts. This generation process yields aligned and context-aware financial news signals, enabling a more rigorous evaluation of news-grounded forecasting models.

Based on this dataset generation pipeline, we introduce XForecast, a benchmark and competition for news-grounded financial forecasting. The task is to predict stock price movements four weeks ahead using both historical price dynamics and paired financial news signals. The dataset covers 100 large-cap stocks from 2019--2023, where models are trained on 2019--2022 and evaluated on 2023.

To better assess models that effectively leverage textual signals, we adopt a weighted hit rate evaluation metric, which measures the correctness of predicted price movement directions while assigning greater importance to instances where news information is expected to have stronger influence. The weights are constructed post hoc by assessing the relevance of news to realized future price movements using a large language model (LLM), allowing us to emphasize instances where news signals are likely to be informative.

We host the XForecast Challenge on Kaggle as part of KDD 2026 Finance Day, providing a standardized evaluation protocol and leaderboard benchmark. Through this benchmark and competition, we aim to encourage research on multimodal financial forecasting and advance methods that more effectively integrate financial TS data with textual information. 

\paragraph{Contributions.}
Our contributions are as follows:
\begin{itemize}[leftmargin=10pt, topsep=2pt, itemsep=1pt]
    \item We introduce XForecast, a benchmark and competition for forecasting stock price movements using both historical financial TS and paired financial news signals.

    \item By integrating stock price data with structured textual information generated using FinTexTS, we provide a way to construct a curated dataset with well-aligned financial text and TS data.

    \item We propose a weighted hit rate metric that places greater emphasis on instances where news information is expected to have stronger influence, with weights estimated through LLM-based post-hoc assessment of news relevance to future price movements.

    \item We host the XForecast Challenge on Kaggle\footnote{\url{https://www.kaggle.com/competitions/xforecast-challenge-kdd}} as part of KDD 2026 Finance Day, providing a standardized evaluation protocol and leaderboard to facilitate research on multimodal financial forecasting.

\end{itemize}

\section{Related Work}

\subsection{Financial Time-Series Forecasting}

Financial TS forecasting has long been studied in quantitative finance and machine learning. 
Traditional approaches rely on statistical models such as ARIMA~\cite{box2015time} and GARCH~\cite{bollerslev1986garch} that model temporal dependencies and volatility dynamics in financial variables. 
Recent advances in deep learning introduce neural architectures including recurrent neural networks~\cite{hochreiter1997lstm}, temporal convolutional networks~\cite{bai2018tcn}, and Transformer-based models~\cite{vaswani2017attention} that capture complex temporal patterns in TS. 
More recent TS-specific architectures further improve long-horizon forecasting performance, including TFT~\cite{lim2021tft}, TimesNet~\cite{wu2023timesnet}, and PatchTST~\cite{nie2023patchtst}. 
These methods typically operate on numerical signals such as historical prices, trading volumes, and technical indicators. 
As a result, the majority of existing financial forecasting research follows a unimodal paradigm in which predictions are derived solely from TS observations.

A limited number of studies attempt to incorporate textual information 
such as financial news, analyst reports, or social media signals for market prediction~\cite{ding2015deep,araci2019finbert}. 
However, many existing datasets provide loosely aligned news streams or focus primarily on sentiment analysis rather than direct forecasting tasks. 
Such settings make it difficult to rigorously evaluate whether textual signals provide predictive information beyond numerical data.

FinTexTS~\cite{lee2026fintexts} is a multimodal dataset generation method that constructs structured pairings between financial texts and time-series signals across multiple contextual levels, including macroeconomic, sector-level, related-company, and company-specific information. 
Such structured alignment enables the study of financial text--time-series interactions. 
However, prior studies using financial text resources largely focus on representation learning or sentiment analysis rather than standardized forecasting benchmarks with clear evaluation protocols.

\subsection{Multimodal Time-Series Forecasting}

Beyond the financial domain, multimodal TS forecasting studies the integration of temporal signals with auxiliary modalities such as text, images, and metadata. 
Recent research explores architectures that combine TS encoders with language models or multimodal representations to incorporate contextual information into temporal prediction tasks, including Time-VLM~\cite{zhong2025timevlm} and TiMi~\cite{lin2026timi}. 
These approaches appear in applications such as weather forecasting and healthcare monitoring, where contextual signals complement numerical measurements.

Several works investigate fusion strategies for combining modalities, including early fusion through input concatenation, cross-modal attention mechanisms, and modular architectures designed for multimodal interaction. 
For example, Time-VLM augments TS forecasting by leveraging representations from vision-language models~\cite{zhong2025timevlm}, while TiMi introduces a multimodal mixture-of-experts architecture that integrates language reasoning with TS transformers~\cite{lin2026timi}. 
Despite these advances, the majority of multimodal forecasting benchmarks focus on domains outside finance, and many existing datasets provide weak or indirect alignment between modalities.

\textbf{Position of the XForecast challenge.}
The dataset used in the XForecast challenge exhibits characteristics of both financial TS forecasting and multimodal forecasting. 
It combines financial TS signals with structured textual information derived from financial news and curated financial texts. 
Specifically, the dataset integrates price dynamics with textual signals generated using FinTexTS~\cite{lee2026fintexts}, enabling models to jointly leverage numerical market signals and news information.

To the best of our knowledge, XForecast represents the first open competition designed to evaluate news-grounded financial forecasting under a standardized benchmark setting. 
By providing aligned multimodal data together with a public leaderboard and evaluation protocol, the challenge aims to facilitate systematic research on integrating financial TS data with textual information.

\section{Challenge Task}
\subsection{Problem Definition}

The goal of the XForecast challenge is to predict future stock price movements using both historical financial TS data and paired textual information. Specifically, participants are asked to forecast the stock price movement four weeks ahead based on information available up to the prediction time. Let $C_{i,t}$ denote the closing price of stock $i$ at time $t$, and let $T_{i,t}$ denote the textual information paired with stock $i$ at time $t$. For each stock and prediction timestamp, models receive historical price observations and associated textual signals up to time $t-4$ weeks. Formally, the model predicts the future closing price as:
\begin{equation}
\begin{aligned}
\hat{C}_{i,t} = f\left(\{C_{i,\tau}\}_{\tau \leq t-4\text{ weeks}}, \{T_{i,\tau}\}_{\tau \leq t-4\text{ weeks}}\right),
\end{aligned}
\end{equation}
where $f(\cdot)$ represents the forecasting model that integrates both numerical TS data and paired textual information.

\subsection{Data Sources}

The XForecast provides a way to construct multimodal datasets by integrating historical stock price data with curated textual signals generated using FinTexTS~\cite{lee2026fintexts}. FinTexTS is a multimodal dataset generation method that produces financial textual signals aligned with financial TS data across multiple contextual levels. By combining price dynamics with structured financial texts, the resulting dataset enables a comprehensive benchmark for evaluating news-grounded financial forecasting models.

\paragraph{FinTexTS.}
FinTexTS is a multimodal dataset generation method for constructing financial text--TS pairs across multiple contextual levels. The method produces semantic-based and multi-level pairings between financial texts and stock market data, capturing macroeconomic context, sector-level information, related-company signals, and company-specific events. This structured pairing framework allows models to incorporate broader financial context beyond firm-specific news and enables more meaningful alignment between textual signals and financial TS dynamics.


\subsection{Evaluation Setting}

The XForecast challenge focuses on large-cap U.S. equities to ensure sufficient market activity and news coverage. Specifically, we select the top 100 companies by market capitalization that were continuously tradable during the period from 2019 to 2023. These companies represent major firms with abundant financial news and stable trading histories, making them suitable for studying news-driven financial forecasting.

We set data from 2019--2022 as the training set, while 2023 is reserved exclusively for evaluation. Models are trained using historical price data and paired textual signals from the training period and are evaluated on their ability to predict stock price movements in the unseen evaluation period.

\subsection{Text Representation}

To incorporate textual information into the forecasting task while ensuring practical feasibility in the Kaggle competition environment, we restrict the use of text to precomputed embedding representations. Direct processing of raw text using large language models during training or inference is not allowed due to the limited computational resources available in the Kaggle runtime environment.

To prevent the use of unapproved pretrained embedding models and ensure a controlled evaluation setting, we restrict text representations to embeddings generated by the following six allowed models:

\begin{itemize}[leftmargin=9pt, topsep=0pt, itemsep=0pt]
\item SBERT~\cite{sbert}
\item LGAI-Embedding-Preview~\cite{lgai}
\item Linq-Embed-Mistral~\cite{linq}
\item llama-embed-nemotron-8b~\cite{nvda}
\item Qwen3-Embedding-8B~\cite{qwen}
\item gemini-embedding-001~\cite{gemini}
\end{itemize}

This design allows participants to explore different strategies for integrating textual signals with financial time-series features while keeping text representations consistent across submissions.

\subsection{Evaluation Metric}

The performance of models in the XForecast challenge is evaluated using a \textbf{weighted hit rate} metric, which measures the accuracy of predicting the direction of stock price movements. For each stock $i$ and evaluation timestamp $t$, let $C_{i,t}$ denote the true closing price and $\hat{C}_{i,t}$ denote the predicted closing price. The predicted direction is considered correct if the predicted price movement direction matches the true movement direction between $t-4$ weeks and $t$. Formally, the directional correctness indicator is defined as
\begin{equation}
\begin{aligned}
H_{i,t}
= 1\Big[
&\operatorname{sign}(\hat{C}_{i,t}-C_{i,t-4\ \mathrm{weeks}}) \\
&= \operatorname{sign}(C_{i,t}-{C}_{i,t-4\ \mathrm{weeks}})
\Big]
\end{aligned}
\end{equation}


Unlike a standard hit rate that assigns equal importance to all instances, we compute a weighted average across evaluation samples:
\begin{equation}
\begin{aligned}
\text{weighted hit rate} =
\frac{1}{\sum_{i,t} w_{i,t}}
\sum_{i,t} w_{i,t} H_{i,t},
\end{aligned}
\end{equation}
where $w_{i,t}$ represents the importance weight assigned to each prediction instance.

The weights are designed to emphasize cases where news information is expected to play a stronger role in explaining subsequent price movements. To compute the weight for each prediction instance, the paired news content is first decomposed into atomic statements. GPT then assigns each statement a positive or negative tag according to its expected impact on the target company. These tags are compared with the realized direction of the company's subsequent price movement, and the instance weight $w_{i,t}$ is determined from the proportion of atomic statements whose tagged impact direction agrees with the realized price direction. Thus, instances for which a larger share of the news statements correctly indicates the subsequent movement receive greater weight in the final metric.

\section{Competition Setup}

\subsection{Competition Rules}

To ensure fair and reproducible evaluation, the XForecast challenge follows several key rules.

First, participants must avoid any form of look-ahead bias. When predicting the closing price at time $t$, models are allowed to use only information available up to $t-4$ weeks. Any use of information within the prediction window is strictly prohibited.

Second, participants are required to submit fully reproducible code. All solutions must be implemented as Kaggle notebooks and must reproduce the submitted predictions within the Kaggle environment. If the submitted results cannot be reproduced by the organizers, the corresponding submission may be disqualified from prize eligibility.

Third, to encourage research on news-grounded forecasting, participants must utilize the provided text embeddings. Submissions that rely solely on numerical price data are excluded from the final ranking.

Finally, to ensure fairness and reproducibility, the use of external datasets or models that are not explicitly permitted is prohibited. Participants must rely only on the data provided in the competition, with limited exceptions for certain time-series foundation models.

\subsection{Prize Structure}

The competition provides monetary awards to the top-performing teams based on the final leaderboard ranking. The total prize pool is \$5,000, distributed as follows:

\begin{itemize}[leftmargin=9pt, topsep=0pt, itemsep=0pt]
\item \textbf{1st Place:} \$3,000
\item \textbf{2nd Place:} \$1,500
\item \textbf{3rd Place:} \$500
\end{itemize}

Prize winners may also be invited to present their solutions at KDD 2026 Finance Day.

\section{Competition Results and Analysis}

\subsection{Participations and Winners}

The challenge ran from June 26 to July 30, 2026. Kaggle recorded 341 submissions during the competition. The top three teams are listed in \Cref{tab:winners}. The first-place solution was submitted by \texttt{finx\_hanyang} from Hanyang University, followed by \texttt{MilkTea} from DeepAuto.ai and \texttt{Kongbyungseung} from Korea University. 

\begin{table}[t]
\caption{Final award recipients.}
\label{tab:winners}
\centering
\small
\setlength{\tabcolsep}{14pt}
\begin{tabular}{cll}
\toprule
Rank & Team & Affiliation \\
\midrule
1 & \texttt{finx\_hanyang} & Hanyang Univ. \\
2 & \texttt{MilkTea} & DeepAuto.ai \\
3 & \texttt{Kongbyungseung} & Korea Univ. \\
\bottomrule
\end{tabular}
\end{table}

\subsection{Representative Solution Strategies}

To characterize the approaches explored by participants, we inspected the code artifacts available for five selected high-ranking teams. This analysis is not intended to represent all 51 teams; rather, it provides a focused view of the methodological diversity among competitive submissions. First, Most inspected solutions were ensembles. Participants combined heterogeneous model families or separated the prediction into market-direction and stock-specific components. Second, cross-sectional information was widely used. Instead of forecasting each stock independently, solutions pooled same-date signals to estimate a common market regime, sector effect, or relative ranking. Third, text was often transformed into compact auxiliary signals---for example, projected embeddings, embedding novelty, pooled news representations, or retrieval features---before fusion with price momentum and volatility features. Finally, both deep architectures and tree-based models remained competitive, suggesting that careful temporal alignment, feature construction, and ensemble design were at least as important as selecting a single model architecture.

\section{Conclusion}

We introduce XForecast, a benchmark and competition for news-grounded financial forecasting that integrates financial time-series data with curated textual signals. By combining structured financial texts generated using FinTexTS with historical stock price data, the dataset enables systematic evaluation of how financial news can be leveraged for stock price forecasting. The challenge focuses on predicting stock price movements four weeks ahead for 100 large-cap companies, with a realistic temporal split between training and evaluation periods. To better assess models that effectively utilize textual signals, we propose a weighted hit rate evaluation metric that emphasizes instances where news information is expected to have a stronger influence.

The XForecast challenge, hosted on Kaggle as part of KDD 2026 Finance Day, attracted 341 submissions, demonstrating substantial interest in news-grounded financial forecasting. The inspected solutions covered diffusion models, inverted Transformers, temporal neural networks, tree ensembles, and foundation-model-based tabular methods, while frequently combining market-wide and stock-specific signals. Beyond leaderboard performance, the post-competition review underscores the importance of precise temporal constraints, unambiguous text-use policies, and reproducible execution environments. We hope this benchmark encourages further research on multimodal financial forecasting and advances methods that effectively integrate numerical market data with textual information.







\bibliography{icml2026_conference}
\bibliographystyle{icml2026}




\end{document}